%% file: main.tex
\documentclass[cameraready]{Interspeech}
\usepackage{bm}
\usepackage{amsmath}
\usepackage{amssymb}
\usepackage{CJKutf8}
\usepackage{booktabs}
\usepackage{multirow}
\usepackage{graphicx}
\usepackage{xcolor}
\usepackage{url}
\usepackage{tikz}
\usepackage{placeins}
\usetikzlibrary{shapes.geometric, arrows.meta, positioning, fit, backgrounds, calc}

\title{Direct Preference Optimization for English-Mandarin Code-Switching Speech Recognition in Audio LLMs}

\author[affiliation={1,2},correspondingauthor=true]{Nguyen Quang}{Trung}
\author[affiliation={1}]{Cheng Yi Lewis}{Won}
\author[affiliation={1}]{Minh Duc}{Pham}
\author[affiliation={1}]{Yingxu}{He}
\author[affiliation={1}]{Shuo}{Sun}
\author[affiliation={1}]{Ai Ti}{Aw}

\address{$^1$Institute for Infocomm Research (I2R), A$^\star$STAR, Singapore\\
$^2$Nanyang Technological University, Singapore}
\email{quangtrung5705@gmail.com}

\makeatletter
\renewcommand\anonemail{\@email\par\vskip 5pt
  {\normalfont\small
  \href{https://ngquangtrung57.github.io/codeswitch-dpo-page/}{\textcolor[HTML]{1A5FB4}{Project Page}} \quad
  \href{https://huggingface.co/collections/ngqtrung/code-switching-dpo-interspeech-2026-6abe22e1354de48b54c0044b}{\textcolor[HTML]{1A5FB4}{Data}}\par}}
\makeatother

\newcommand\blfootnote[1]{%
  \begingroup
  \renewcommand\thefootnote{}\footnote{#1}%
  \addtocounter{footnote}{-1}%
  \endgroup
}

\begin{document}

\maketitle
\blfootnote{Accepted to Interspeech 2026. DOI: \href{https://doi.org/10.21437/Interspeech.2026-110}{10.21437/Interspeech.2026-110}.}
\blfootnote{Work done during Nguyen Quang Trung's internship at Institute for Infocomm Research (I2R), A$^\star$STAR.}

\begin{abstract}
Audio large language models (Audio LLMs) exhibit systematic failures in transcribing code-switching speech despite strong multilingual capabilities. Focusing on English-Mandarin, we identify three failure modes: language omission, translation-instead-of-transcription, and hallucination. We apply Direct Preference Optimization (DPO) to align models, constructing preference pairs in which chosen responses preserve mixed-language content while rejected responses mimic failure patterns. Training three Audio LLMs on 100K pairs (570 hours), we observe consistent behavioral shifts: models learn to preserve language composition rather than translating when prompted for transcription. This alignment yields MER reductions up to 89.6\% (in-distribution) and 20.0\% (out-of-distribution). Our findings suggest DPO can effectively elicit correct code-switching transcription behavior from multilingual Audio LLMs.
\end{abstract}

\section{Introduction}
\label{sec:intro}

Audio large language models (Audio LLMs) extend large language models with the ability to process and understand audio inputs alongside text, enabling tasks such as speech recognition, audio captioning, and spoken dialogue~\cite{radford2023robust,chu2024qwen2,tang2023salmonn}. Since Whisper~\cite{radford2023robust} established the foundation through large-scale weak supervision, numerous models have emerged with strong multilingual competence: Qwen2-Audio~\cite{chu2024qwen2} and the Qwen-Omni series~\cite{xu2025qwen2, Qwen3-Omni} support multiple languages, Phi-4 Multimodal~\cite{abouelenin2025phi} demonstrates multilingual performance through its Mixture-of-LoRAs architecture, and MERaLiON~\cite{he-etal-2025-meralion} is specifically designed for Southeast Asian multilingual contexts. These advances suggest that modern Audio LLMs possess multilingual proficiency, as evidenced by state-of-the-art performance on benchmarks such as Common Voice~\cite{ardila2020common} and FLEURS~\cite{conneau2023fleurs}.

However, this multilingual capability does not automatically extend to code-switching, the practice of alternating between languages within a conversation or utterance, which is prevalent in multilingual communities worldwide. We focus on English-Mandarin code-switching because it is one of the most widely studied language pairs in Southeast Asia, where the SEAME corpus~\cite{lyu10_interspeech,zeng2018seame} serves as a standard benchmark. Despite the multilingual proficiency of Audio LLMs, even models like MERaLiON, which incorporate extensive code-switching data during supervised fine-tuning, exhibit systematic failures when transcribing code-switching speech. Through our analysis, we identify three distinct failure modes: (1) \textbf{language omission}, where the model outputs only one language while dropping the other; (2) \textbf{translation-instead-of-transcription}, where the model translates mixed-language content into a single language rather than preserving the original; and (3) \textbf{hallucination}, where the model generates repeated or fabricated content.

Prior work has explored various directions to address code-switching in Automatic Speech Recognition (ASR). Early systems relied on hybrid pipelines such as phone merging~\cite{vu2012first} and factored language models~\cite{adel2013combination}. More recent approaches focus on reducing reliance on naturally occurring code-switching data via audio concatenation~\cite{hussein2024collage,nguyen2025noreal}, adapting foundation models through encoding refinement with language-aware decoding~\cite{zhao2025whispercs,liu2023reducing,liu2024interactive}, and employing Mixture-of-Experts architectures for language-specialized processing~\cite{zhang2025moe,ye2024scmoe}. However, none explicitly target the behavioral alignment of Audio LLMs for code-switching transcription.

We hypothesize that Audio LLMs possess the latent ability to produce correct code-switching transcriptions, and this behavior can be elicited through preference optimization. To test this hypothesis, we turn to Direct Preference Optimization (DPO)~\cite{rafailov2023dpo}, a common direct alignment algorithm for large language models. Within the speech domain, SpeechAlign~\cite{zhang2024speechalign} first demonstrated DPO's effectiveness for aligning codec language models. Notably, Qwen2-Audio~\cite{chu2024qwen2} already incorporates DPO in its training pipeline. However, to our knowledge, no existing work applies DPO specifically to address code-switching transcription capability in Audio LLMs. Therefore, we investigate DPO as a potential approach: given models with demonstrated multilingual capability, we ask whether DPO can elicit correct code-switching transcription behavior.

To implement this approach, we construct DPO training pairs by pairing ground-truth code-switching transcriptions (chosen) with synthetically generated flawed transcriptions (rejected) that mimic the observed failure modes. Using approximately 100K preference pairs ($\sim$570 hours) derived from natural and synthetic code-switching data, we train three Audio LLMs -- MERaLiON-2-3B~\cite{he-etal-2025-meralion}, Phi-4-multimodal-instruct~\cite{abouelenin2025phi}, and Qwen2-Audio-7B-Instruct~\cite{chu2024qwen2} -- and evaluate on both in-distribution and out-of-distribution benchmarks, including SEAME dev\_man and dev\_sge~\cite{zeng2018seame}.

In summary, our contributions are as follows:
\begin{itemize}
    \item We identify three systematic failure modes in English-Mandarin code-switching transcription exhibited by state-of-the-art multilingual Audio LLMs.
    \item We propose a DPO approach to construct preference pairs that contrast correct code-switching transcriptions with failure-mimicking alternatives.
    \item We demonstrate consistent improvements across three Audio LLM architectures on English-Mandarin benchmarks, achieving relative MER reductions of up to 20.0\% (out-of-distribution) and 89.6\% (in-distribution).
\end{itemize}

\section{Method}

Our approach consists of two main components: (1) constructing preference pairs where ground-truth code-switching transcriptions serve as chosen responses and LLM-generated flawed transcriptions serve as rejected responses, and (2) applying DPO to align model transcription behavior toward the correct code-switching output. Figure~\ref{fig:method} provides an overview of this pipeline, while Table~\ref{tab:failure_modes} illustrates the three failure modes introduced in Section~\ref{sec:intro} with concrete examples.

\input{figures/method_overview}

\begin{table}[th]
  \caption{Three failure modes observed in Audio LLMs on code-switching audio inputs when prompted for transcription}
  \label{tab:failure_modes}
  \centering
  \footnotesize
  \begin{tabular}{p{1.5cm}p{1.8cm}p{2.7cm}}
    \toprule
    \textbf{Failure Mode} & \textbf{Description} & \textbf{Example} \\
    \midrule
    Language Omission & Drops one language entirely & GT: ``\begin{CJK}{UTF8}{gbsn}我住\end{CJK} temasek poly \begin{CJK}{UTF8}{gbsn}那边\end{CJK}'' \newline Out: ``\begin{CJK}{UTF8}{gbsn}我住那边\end{CJK}'' \\
    \addlinespace
    Translation & Translates instead of transcribes & GT: ``\begin{CJK}{UTF8}{gbsn}我们都应该\end{CJK} pursue a healthy lifestyle'' \newline Out: ``\begin{CJK}{UTF8}{gbsn}我们都应该追求健康的生活方式\end{CJK}'' \\
    \addlinespace
    Hallucination & Repeated or fabricated content & GT: ``\begin{CJK}{UTF8}{gbsn}我们二月多有\end{CJK} valentine's day'' \newline Out: ``ah month...'' ($\times$250) \\
    \bottomrule
  \end{tabular}
\end{table}

\subsection{DPO for code-switching alignment}

We apply DPO to align Audio LLM output behavior for code-switching transcription. Specifically, given an input $\mathbf{x}$ (an audio clip and a prompt asking for transcription), a preferred response $\mathbf{y}_c$ (ground-truth transcription), and a dispreferred response $\mathbf{y}_r$ (flawed transcription), DPO optimizes the policy $\pi_\theta$ to increase the likelihood of generating $\mathbf{y}_c$ while decreasing the likelihood of generating $\mathbf{y}_r$, directly without explicit reward modeling:
\begin{align}
\mathcal{L}_{\text{DPO}} = -\mathbb{E} \Big[ \log \sigma \big( \beta \log \frac{\pi_\theta(\mathbf{y}_c | \mathbf{x})}{\pi_{\text{ref}}(\mathbf{y}_c | \mathbf{x})} - \beta \log \frac{\pi_\theta(\mathbf{y}_r | \mathbf{x})}{\pi_{\text{ref}}(\mathbf{y}_r | \mathbf{x})} \big) \Big]
\label{eq:dpo}
\end{align}
where $\pi_\theta$ is the active policy being trained, $\pi_{\text{ref}}$ is the reference policy (frozen base model), $\sigma$ is the sigmoid function, $\beta$ controls preference strength, $\mathbf{y}_c$ (chosen) is the ground-truth code-switching transcription, and $\mathbf{y}_r$ (rejected) is the response that mimics the observed failure modes.

\subsection{Training data construction}

\subsubsection{Rejected sample generation}

To create the rejected samples, we use Qwen3-32B~\cite{yang2025qwen3} to transform ground-truth transcriptions into flawed versions. We employ two complementary strategies, both targeting translation-based failure modes:

\noindent\textbf{Global Translation (80\%):} This strategy translates all content from one language to the other (all Chinese$\rightarrow$English or all English$\rightarrow$Chinese), thereby mimicking translation-instead-of-transcription failures.

\noindent\textbf{Partial Translation (20\%):} In contrast, this strategy translates only specific short spans within the utterance, mimicking partial language omission where isolated segments are incorrectly rendered in the wrong language.

We chose the 80/20 ratio based on the higher frequency of full translation errors observed in baseline models. Table~\ref{tab:rejected_examples} illustrates both strategies with examples. Even though both strategies focus on translation-based rejected samples, and we do not explicitly generate rejected samples representing content dropping or hallucination (repetition/fabrication), we show in Section~\ref{sec:results} that DPO training reduces all three failure modes. This is likely because models trained to preserve language composition also develop more stable generation patterns.

\begin{table}[th]
  \caption{Examples of rejected sample generation strategies}
  \label{tab:rejected_examples}
  \centering
  \small
  \begin{tabular}{p{0.95\columnwidth}}
    \toprule
    \textbf{Global Translation (EN$\rightarrow$CN)} \\
    \midrule
    \textbf{Chosen:} What grade are you? \begin{CJK}{UTF8}{gbsn}真的很好哎，真的前途无限呀。\end{CJK} \\
    \textbf{Rejected:} \begin{CJK}{UTF8}{gbsn}你几年级？真的很好哎，真的前途无限呀。\end{CJK} \\
    \textit{All English translated to Chinese.} \\
    \midrule
    \textbf{Partial Translation (EN$\rightarrow$CN)} \\
    \midrule
    \textbf{Chosen:} \begin{CJK}{UTF8}{gbsn}基本每天就是做题刷题。\end{CJK} It's so boring and dull. \\
    \textbf{Rejected:} \begin{CJK}{UTF8}{gbsn}基本每天就是做题刷题。\end{CJK} It's so \begin{CJK}{UTF8}{gbsn}无聊\end{CJK} and dull. \\
    \textit{Only ``boring'' translated; rest preserved.} \\
    \bottomrule
  \end{tabular}
\end{table}

\subsubsection{Data sources}

We construct DPO training pairs from two complementary datasets as summarized in Table~\ref{tab:data_sources}.

\begin{table}[th]
  \caption{Training data composition for DPO}
  \label{tab:data_sources}
  \centering
  \begin{tabular}{lrr}
    \toprule
    \textbf{Source} & \textbf{Hours} & \textbf{Pairs} \\
    \midrule
    CS-Dialogue~\cite{zhou2025csdialogue} & 77.3 & 13,795 \\
    EMILIA~\cite{he2024emilia} & 489.5 & 86,971 \\
    \midrule
    \textbf{Total} & \textbf{566.8} & \textbf{100,766} \\
    \bottomrule
  \end{tabular}
\end{table}

\noindent\textbf{CS-Dialogue~\cite{zhou2025csdialogue}:} This contains spontaneous Mandarin-English code-switching dialogues from 200 speakers, where each utterance is tagged as English-only (EN), Chinese-only (CN), or code-switched (MIX). From this data, we construct segments in two ways: first, by grouping consecutive MIX-tagged utterances containing natural intra-sentential code-switching; and second, by concatenating EN and CN utterances from the same conversation to create inter-sentential code-switching. Together, these approaches combine authentic spontaneous code-switching with controlled cross-utterance language mixing within coherent conversational contexts.

\noindent\textbf{EMILIA~\cite{he2024emilia}:} To complement CS-Dialogue with additional scale and diversity, we create synthetic code-switching samples by concatenating English and Chinese clips from the EMILIA corpus. Each segment is constructed by randomly sampling clips from both languages and concatenating them, producing inter-sentential code-switching audio at scale.

\section{Experimental setup}
\label{sec:experiments}

\subsection{Models}

To demonstrate the generalizability of our approach, we experiment with three multilingual Audio LLMs, which are already trained in both English and Mandarin. Table~\ref{tab:models} summarizes the training configurations.

\begin{table}[t]
    \caption{DPO training configurations. We tuned $\beta$ per model based on each architecture's sensitivity to the strength of preference optimization: lower $\beta$ enables stronger updates, while higher $\beta$ produces more conservative changes. All hyperparameters were selected through validation-based tuning on held-out splits of the training data.}
  \label{tab:models}
  \centering
  \small
   \resizebox{1\columnwidth}{!}{\begin{tabular}{lccrrc}
    \toprule
    \textbf{Model} & \textbf{\#P} & \textbf{Method} & $\bm{\beta}$ & \textbf{LR} & \textbf{BS} \\
    \midrule
    MERaLiON-2-3B & 3B & Full & 0.5 & 1e-6 & 256 \\
    Phi-4-MM & 6B & Full & 0.05 & 5e-6 & 256 \\
    Qwen2-Audio-7B-IT & 7B & LoRA & 0.3 & 3e-5 & 64 \\
    \bottomrule
  \end{tabular}}
\end{table}

\textbf{MERaLiON-2-3B}~\cite{he-etal-2025-meralion} is specifically designed for Southeast Asian multilingual speech and includes extensive code-switching data in its supervised fine-tuning stage.

\textbf{Phi-4-multimodal-instruct}~\cite{abouelenin2025phi} is a general-purpose multimodal model with strong multilingual capability, including English and Mandarin.

\textbf{Qwen2-Audio-7B-Instruct}~\cite{chu2024qwen2} is a foundational Audio LLM with state-of-the-art performance across multiple audio benchmarks. For this model, we apply LoRA adaptation with rank 256 targeting all attention and MLP modules, because preliminary experiments with full fine-tuning across multiple hyperparameter configurations consistently produced degraded outputs with repetitive tokens and severe hallucinations. Thus, LoRA preserves more stable generation behavior.

All three models were trained for one epoch on 8 H100 GPUs.

\subsection{Prompt diversity}

Audio LLMs require both audio input and a text prompt to perform transcription. To prevent overfitting to a single prompt template during training, we use 20 English and 20 Chinese prompts, all requesting transcription but with varied phrasing. Examples include: ``\textit{Please transcribe the speech in this audio file.}'', ``\textit{Can you transcribe this audio for me?}'', ``\begin{CJK}{UTF8}{gbsn}\textit{请帮我转写这段音频。}\end{CJK}'' (Please transcribe this audio), and ``\begin{CJK}{UTF8}{gbsn}\textit{这段音频里在说什么？}\end{CJK}'' (What is being said in this audio?). During training, prompts are randomly sampled from this pool; during evaluation, we use a fixed prompt: ``\textit{Please transcribe this speech.}'', which is a common prompt and not included in the training pool.

\subsection{Evaluation benchmarks}

We evaluate on four benchmarks (Table~\ref{tab:benchmarks}), including both in-distribution and out-of-distribution test sets to assess generalization.

\begin{table}[t]
  \caption{Evaluation benchmarks for English-Mandarin code-switching ASR.}
  \label{tab:benchmarks}
  \centering
  \begin{tabular}{lrr}
    \toprule
    \textbf{Benchmark} & \textbf{Samples} & \textbf{Hours} \\
    \midrule
    SEAME dev\_man~\cite{zeng2018seame} & 2,610 & 2.0 \\
    SEAME dev\_sge~\cite{zeng2018seame} & 3,222 & 2.5 \\
    EMILIA-test~\cite{he2024emilia} & 1,000 & 5.6 \\
    CS-Dialogue-test~\cite{zhou2025csdialogue} & 359 & 2.1 \\
    \bottomrule
  \end{tabular}
\end{table}

\textbf{SEAME}~\cite{lyu10_interspeech} is a standard English-Mandarin code-switching corpus collected from conversational speech in Singapore and Malaysia. We evaluate on its dev\_man and dev\_sge splits~\cite{zeng2018seame} (2,610 and 3,222 utterances, respectively), which serve as our out-of-distribution test sets, since no SEAME data appears in the training set.

\textbf{EMILIA-test} and \textbf{CS-Dialogue-test}, in contrast, are held-out portions of the training and validation data sources and thus represent in-distribution evaluation.

\subsection{Evaluation metric}

We use Mixed Error Rate (MER), a standard metric for code-switching ASR evaluation. MER applies character-level tokenization to Chinese text and word-level tokenization to English text, respecting the natural linguistic structure of both languages. To ensure accurate MER calculation, all text is lowercased, and punctuation is removed before evaluation. Moreover, we apply model-specific output normalization. For example, Qwen2-Audio-7B-Instruct typically outputs ``The original content of this audio is: [transcription]''. We filter such patterns to extract only the transcription content, ensuring a fair comparison. Lower MER indicates better performance.

\section{Results}
\label{sec:results}

\subsection{Quantitative analysis}

Table~\ref{tab:main_results} presents MER scores across all models and benchmarks. These results show that DPO consistently improves code-switching transcription performance across all configurations, though the magnitude of improvement varies considerably by model and benchmark.

\begin{table}[t]
  \caption{Main results (MER \%, lower is better). All models show consistent improvement after DPO training.}
  \label{tab:main_results}
  \centering
  \small
  \resizebox{1\columnwidth}{!}{
  \begin{tabular}{llrrr}
    \toprule
    \textbf{Model} & \textbf{Benchmark} & \textbf{Base} & \textbf{DPO} & \textbf{$\Delta$Rel} \\
    \midrule
    \multirow{4}{*}{MERaLiON-2-3B}
      & SEAME dev\_sge & 32.38 & 31.75 & -2.0\% \\
      & SEAME dev\_man & 25.79 & 25.61 & -0.7\% \\
      & EMILIA & 32.01 & 30.41 & -5.0\% \\
      & CS-Dialogue & 25.41 & \textbf{22.58} & \textbf{-11.1\%} \\
    \midrule
    \multirow{4}{*}{\shortstack[l]{Phi-4-multimodal\\-instruct}}
      & SEAME dev\_sge & 69.97 & 61.09 & -12.7\% \\
      & SEAME dev\_man & 51.97 & 46.63 & -10.3\% \\
      & EMILIA & 70.98 & \textbf{7.38} & \textbf{-89.6\%} \\
      & CS-Dialogue & 49.61 & 10.65 & -78.5\% \\
    \midrule
    \multirow{4}{*}{\shortstack[l]{Qwen2-Audio\\-7B-Instruct}}
      & SEAME dev\_sge & 95.11 & 85.52 & -10.1\% \\
      & SEAME dev\_man & 72.89 & \textbf{58.30} & \textbf{-20.0\%} \\
      & EMILIA & 44.70 & 42.08 & -5.9\% \\
      & CS-Dialogue & 38.91 & 31.40 & -19.3\% \\
    \bottomrule
  \end{tabular}
  }
\end{table}

\textbf{MERaLiON-2-3B} shows modest SEAME improvements (0.7--2.0\%), which reflects that this model already incorporates extensive code-switching data in its supervised training, leaving limited room for further gains. Nevertheless, DPO still provides an 11.1\% relative improvement on in-distribution CS-Dialogue data.

\textbf{Phi-4-multimodal-instruct}, on the other hand, shows dramatic improvement: MER drops from 70.98\% to 7.38\% on EMILIA (89.6\% relative reduction). We attribute this large gain to the model's limited exposure to code-switching during its training process, which causes the baseline to frequently translate mixed-language content to a single language or produce severe repetition when asked to transcribe. Hence, a single epoch of DPO effectively elicits correct code-switching transcription behavior.

\textbf{Qwen2-Audio-7B-Instruct} demonstrates substantial improvement on SEAME dev\_man (20.0\% relative), alongside consistent gains on in-distribution benchmarks (5.9--19.3\%).

\subsection{Qualitative analysis}

Beyond aggregate MER scores, our primary goal is \emph{behavioral alignment} for code-switching transcription: when asked to transcribe mixed-language audio, the model should preserve the original language composition rather than translating to a single language, dropping one language, or producing repetitive content. In manual inspection of model outputs, we frequently observe that DPO shifts generation toward the desired behavior: models become more likely to preserve the mixed-language pattern and produce more stable transcriptions. To illustrate these behavioral changes, Table~\ref{tab:examples} shows representative corrections after DPO training.

\begin{table}[t]
  \caption{Qualitative examples showing DPO corrections.}
  \label{tab:examples}
  \centering
  \footnotesize
  \begin{tabular}{p{0.9\columnwidth}}
    \toprule
    \textbf{Example 1: Translation-Instead-of-Transcription (Qwen2-Audio-7B-Instruct)} \\
    \midrule
    \textbf{GT:} \begin{CJK}{UTF8}{gbsn}我们都应该\end{CJK} pursue a healthy lifestyle \\
    \textbf{Base:} \begin{CJK}{UTF8}{gbsn}我们都应该追求健康的生活方式\end{CJK} \\
    \textbf{DPO:} \begin{CJK}{UTF8}{gbsn}我们都应该\end{CJK} pursue a healthy lifestyle \\
    \textit{MER: 100\% $\rightarrow$ 0\%} \\
    \midrule
    \textbf{Example 2: Hallucination (Phi-4-multimodal-instruct)} \\
    \midrule
    \textbf{GT:} \begin{CJK}{UTF8}{gbsn}我们二月多有\end{CJK} valentine's day \\
    \textbf{Base:} ah month ah month ah month... ($\times$250) \\
    \textbf{DPO:} \begin{CJK}{UTF8}{gbsn}二月多有\end{CJK} Valentine's Day \\
    \textit{MER: 56.89\% $\rightarrow$ 0.33\%} \\
    \midrule
    \textbf{Example 3: Language Omission (MERaLiON-2-3B)} \\
    \midrule
    \textbf{GT:} \begin{CJK}{UTF8}{gbsn}我住\end{CJK} temasek poly \begin{CJK}{UTF8}{gbsn}那边\end{CJK} \\
    \textbf{Base:} \begin{CJK}{UTF8}{gbsn}我住达马士科波利那边\end{CJK} (transliterated) \\
    \textbf{DPO:} \begin{CJK}{UTF8}{gbsn}我住\end{CJK} tamasek poly \begin{CJK}{UTF8}{gbsn}那边\end{CJK} \\
    \textit{MER: 100\% $\rightarrow$ 17\%} \\
    \bottomrule
  \end{tabular}
\end{table}

\subsection{Analysis}

Taken together, these results show that DPO consistently shifts transcription behavior toward the desired code-switching output across three model families. These qualitative corrections support our hypothesis that the ability to produce accurate code-switching transcriptions is often latent in multilingual Audio LLMs but is not reliably expressed when prompted to transcribe. DPO thus provides a lightweight mechanism to elicit the intended code-switching behavior from such models through preference pairs. Because our goal is behavioral alignment, we consider these qualitative shifts in code-switching awareness to be a primary outcome alongside MER.

\section{Discussion}
\label{sec:discussion}

\textbf{Limitations.} While our results are encouraging, several aspects of our approach present opportunities for refinement.
\textit{First,} we focus exclusively on English-Mandarin code-switching; generalization to other language pairs remains untested.
\textit{Second,} our rejected samples are synthetic transformations rather than samples drawn from the model's actual outputs. This provides controllability and scalability, but it may introduce distributional shifts relative to real failure modes.
\textit{Third,} we employ vanilla DPO without algorithmic modifications; recent variants such as SimPO~\cite{meng2024simpo}, mDPO~\cite{wang2024mdpo}, or iterative refinement approaches could potentially yield further improvements.
\textit{Fourth,} our rejection strategy focuses primarily on translation-based failures; explicit generation of hallucination and content-dropping examples might strengthen alignment for those specific modes.

\section{Conclusion}
\label{sec:conclusion}

In this work, we applied Direct Preference Optimization to address English-Mandarin code-switching failures in three Audio LLMs: MERaLiON-2-3B, Phi-4-multimodal-instruct, and Qwen2-Audio-7B-Instruct.
Starting from the observation that these models exhibit systematic failure modes -- language omission, translation-instead-of-transcription, and hallucination -- we constructed approximately 100K DPO training pairs ($\sim$570 hours) using LLM-generated rejected samples that mimic these failures.

Our experiments demonstrate that DPO training yields consistent improvements across all models and benchmarks, with relative MER reductions of up to 89.6\% on in-distribution data (Phi-4 on EMILIA) and 20.0\% on out-of-distribution data (Qwen2-Audio on SEAME dev\_man).
Furthermore, qualitative analysis confirms that DPO effectively corrects all three failure modes, enabling models to produce verbatim mixed-language transcriptions.

To our knowledge, this is among the first demonstrations that Direct Preference Optimization can elicit correct code-switching transcription behavior from multilingual Audio LLMs, and it offers a viable mechanism for aligning models that already possess multilingual capability. We hope this work opens avenues for extending to other language pairs, exploring alternative preference optimization algorithms, and advancing broader Audio LLM architectures.

\bibliographystyle{IEEEtran}
\bibliography{references_dpo_codeswitching}

\end{document}

%% file: figures/method_overview.tex
\begin{figure*}[t]
\centering
\resizebox{\textwidth}{!}{%
\begin{tikzpicture}[
    node distance=0.8cm and 1.0cm,
    box/.style={
        rectangle, rounded corners=4pt, minimum height=1.1cm,
        draw=#1!70!black, fill=#1!10, font=\large, align=center, line width=1pt,
        minimum width=2.5cm
    },
    inputbox/.style={box=gray},
    chosenbox/.style={box=green, minimum width=3.5cm},
    rejectedbox/.style={box=red, minimum width=3.5cm},
    processbox/.style={box=blue},
    llmbox/.style={box=orange},
    pairbox/.style={box=violet, minimum width=2.8cm, minimum height=1.5cm},
    myarrow/.style={-{Stealth[length=3mm, width=2mm]}, line width=1.2pt, #1},
    extext/.style={font=\normalsize, text=gray!70},
]


\node[inputbox] (audio) {\textbf{Audio Input}\\\textbf{Code-Switching}};

\node[inputbox, right=1.5cm of audio] (gt) {\textbf{Ground Truth}\\\textbf{Transcription}};

\node[llmbox, right=1.5cm of gt] (llm) {\textbf{Qwen3-32B}\\\textbf{Rejection Generator}};

\node[chosenbox, right=2.5cm of llm, yshift=0.9cm] (chosen) {
    \textbf{Chosen} $\mathbf{y}_c$\\
    {\large\begin{CJK}{UTF8}{gbsn}我住\end{CJK} temasek poly \begin{CJK}{UTF8}{gbsn}那边\end{CJK}}
};

\node[rejectedbox, right=2.5cm of llm, yshift=-0.9cm] (rejected) {
    \textbf{Rejected} $\mathbf{y}_r$\\
    {\large ``I live temasek poly there''}
};

\node[pairbox, right=1.8cm of chosen, yshift=-0.9cm] (pair) {
    \textbf{Preference}\\
    \textbf{Pair}\\
    {\large$(\mathbf{x}, \mathbf{y}_c, \mathbf{y}_r)$}
};

\node[processbox, right=1.5cm of pair] (dpo) {\textbf{DPO}\\\textbf{Training}};

\node[processbox, right=1.5cm of dpo, fill=purple!10, draw=purple!70!black] (output) {\textbf{Aligned}\\\textbf{Audio LLM}};


\draw[myarrow=gray!60] (audio) -- (gt);

\draw[myarrow=gray!60] (gt) -- (llm);

\draw[myarrow=green!60!black] (gt.east) -- ++(0.4,0) |- (chosen.west);

\draw[myarrow=red!60!black] (llm.east) -- ++(0.4,0) |- (rejected.west);

\draw[myarrow=green!60!black] (chosen.east) -- ++(0.4,0) |- ([yshift=0.25cm]pair.west);

\draw[myarrow=red!60!black] (rejected.east) -- ++(0.4,0) |- ([yshift=-0.25cm]pair.west);

\draw[myarrow=blue!60] (pair) -- (dpo);

\draw[myarrow=purple!60] (dpo) -- (output);


\node[below=0.4cm of llm, font=\large, align=center, text=red!70!black] {
    Global Translation (80\%)\\[-1pt]
    Partial Translation (20\%)
};

\node[right=-0.1cm of chosen.north east, text=green!60!black, font=\normalsize] {\checkmark};
\node[right=-0.1cm of rejected.north east, text=red!60!black, font=\normalsize] {$\times$};

\end{tikzpicture}%
}
\caption{Overview of DPO training for code-switching alignment. Ground-truth transcriptions serve as chosen responses ($\mathbf{y}_c$), while an LLM generates rejected responses ($\mathbf{y}_r$) that mimic failure modes via Global Translation (full) and Partial Translation (spans only). DPO trains the model to prefer verbatim code-switching output.}
\label{fig:method}
\end{figure*}
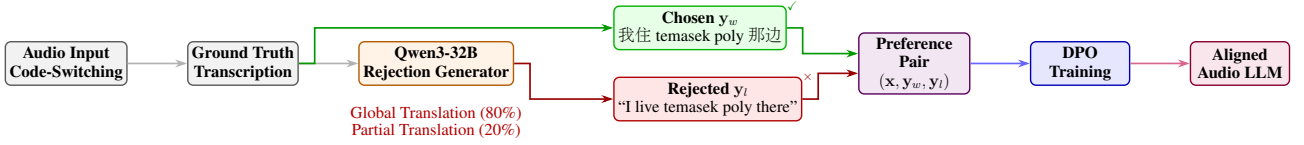